\documentclass[10pt,twocolumn,letterpaper]{article}

\usepackage{iccv}
\usepackage{times}
\usepackage{epsfig}
\usepackage{graphicx}
\usepackage{amsmath}
\usepackage{amssymb}

\usepackage{booktabs}
\usepackage{xcolor}
\usepackage{amsmath, bm, subfigure, epstopdf, url, pifont, overpic, cases}
\usepackage{latexsym, amssymb, bbding, multirow, makecell,  diagbox, enumitem, caption}
\usepackage{array, enumitem, soul, algorithm, algpseudocode}

\usepackage{arydshln}

\usepackage{multirow, multicol}
\usepackage{adjustbox}

\newlength \g

\usepackage[pagebackref=true,breaklinks=true,letterpaper=true,colorlinks,bookmarks=false]{hyperref}

\iccvfinalcopy %

\def\iccvPaperID{10} %

\begin{document}

\title{Towards 3D-Aware Video Diffusion Models: Render-Free Human Motion Control with Mesh Tokenization}

\author{Jingyun Liang$^{1,2}$ ~ Min Wei$^{1,2}$ ~ Shikai Li$^{1,2}$ ~ Yizeng Han$^{1,2}$  \\ Hangjie Yuan$^{1,2,4}$  ~  Lei Sun$^{3}$ ~ Weihua Chen$^{1,2}$ ~ Fan Wang$^{1,2}$  ~ \\
$^{1}$DAMO Academy, Alibaba Group ~~~~ $^{2}$Hupan Lab ~~~~ $^{3}$Zhejiang University ~~~~ $^{4}$INSAIT \\
{\tt\small }\url{https://jingyunliang.github.io/MeshToken}
}

\maketitle

\begin{abstract}
Diffusion models have shown remarkable success in video generation. However, whether such models are truly aware of the 3D structure underlying visual observations, rather than simply reproducing plausible 2D projections, remains an open question. In this work, we investigate this question through human motion control, a task that requires precise modelling of 3D human geometry, motion, camera viewpoint, and scene context. Unlike prior methods that rely on rendered 2D motion guidance videos, we propose a render-free framework that conditions video generation directly on compressed 3D human mesh tokens. This representation preserves full 3D geometric information while enabling a unified token-based generation pipeline that processes video tokens jointly with motion tokens in a DiT-based architecture. This design requires the model to reason jointly about appearance, 3D structure, and camera viewpoint during video generation. Experimental results demonstrate strong performance on human motion control benchmarks, while reducing artifacts induced by view-dependent 2D guidance and trajectory-pose mismatches during editing. These findings suggest that video diffusion models, when equipped with mesh tokenization, can better capture complex 3D human structures and their interactions with the surrounding environment.

\end{abstract}

\section{Introduction}
Recent advances in diffusion models have revolutionized video generation~\cite{blattmann2023stable, hu2024animate, rombach2022high, wang2025wan, hu2025animate, jiang2025vace}. With their strong ability to generate plausible object motion and physical dynamics, video diffusion models (VDMs) have shown impressive potential for modeling dynamic visual scenes~\cite{chen2024diffusion, rigter2024avid, agarwal2025cosmos, huang2025vid2world}. This naturally raises a fundamental question: \textbf{Are video diffusion models truly aware of the underlying 3D structure of the visual world, or do they simply reproduce plausible 2D observations?}

Answering this question is challenging. A model that produces visually realistic videos may still rely primarily on statistical regularities in 2D image sequences, without forming a robust understanding of the underlying 3D structure. This ambiguity is particularly evident in controllable human video generation, where accurate synthesis requires jointly modeling articulated 3D body geometry, motion dynamics, camera viewpoint, and scene context. \textbf{Human motion control therefore provides a suitable test bed for probing 3D awareness}: to generate a realistic video following a target motion, the model must not only preserve appearance, but also correctly interpret body structure, motion trajectory, and their projection under different camera views.

However, most existing approaches to human motion control operate in the 2D space by projecting 3D motion into rendered guidance signals, such as pose maps, skeleton videos, or other rasterized control signals. While effective in practice, these rendering-based representations are inherently view-dependent and often discard critical 3D geometric information, including occluded or back-facing surfaces. As a result, the model may learn to mimic the appearance of guidance frames, effectively reducing the task to video-to-video translation rather than reasoning about the underlying 3D motion itself. This limitation becomes even more pronounced in editing scenarios, where imperfect alignment between trajectory, pose, and camera viewpoint can easily introduce structural artifacts such as floating feet, penetration, and motion inconsistency. Therefore, existing rendering-based paradigms are not well suited to studying whether video diffusion models can truly become 3D-aware.

To address the above limitation, we propose a render-free framework for human motion control that directly conditions video generation on compressed 3D human mesh tokens. More specifically, we first represent human motion using the parametric 3D human model SMPL~\cite{loper2023smpl}, and decompose it into trajectory and 3D body pose mesh sequences. We then compress the mesh sequence into compact latent tokens using a vector-quantized variational autoencoder (VQ-VAE)~\cite{esser2021taming, fiche2024vq}, and combine them with trajectory embeddings to form motion tokens. These motion tokens are injected into a DiT-based video generation model through cross-attention, yielding a unified token-based framework that jointly models visual appearance and 3D motion.

Our formulation provides a more direct way to probe 3D awareness in video diffusion models. By removing the rendering bottleneck, the motion condition is no longer tied to a specific camera view, which reduces the possibility that the model solves the task through view-dependent 2D shortcuts. The model therefore needs to infer how the same underlying 3D motion should appear under different viewpoints. To faithfully generate the desired video, the model further needs to jointly reason about articulated 3D human motion, camera viewpoint, scene context, and human-environment interaction. This enables us to assess the model's 3D awareness through the generated videos.

Moreover, for the human motion control task itself, compressing 3D human meshes into compact tokens preserves full 3D geometric fidelity, retaining occluded geometry and improving multi-view and temporal consistency. As a structured token representation, it also integrates naturally with token-based video generation pipelines. By disentangling motion into trajectory and body pose, our framework offers a more flexible and robust interface for motion editing than rendering-based methods, and better supports compositional control over different motion factors. In contrast, rendering-based methods often suffer from motion artifacts such as feet floating, penetration, and sliding due to imperfect matching between trajectory and body pose during editing.

The contributions of this paper are summarized as follows:
\begin{itemize}
    \item We study the 3D awareness of video diffusion models through the lens of human motion control, and propose the first render-free framework that directly conditions video generation on 3D human mesh tokens, avoiding the information loss and view-dependent bias of rendering-based approaches. This opens a new perspective for 3D-aware video generation and control.
    \item We introduce a mesh tokenization pipeline that decomposes human motion into trajectory and body pose, and compresses 3D meshes into compact motion tokens. This yields a geometrically faithful and disentangled motion representation within a unified token-based DiT architecture.
    \item Extensive experiments demonstrate that our method achieves strong performance in human motion control and exhibits stronger robustness under viewpoint changes and motion editing. These results suggest that video diffusion models are capable of capturing richer 3D structure when provided with appropriate 3D representations.
\end{itemize}

\begin{figure*}[t]
    \centering
    \includegraphics[width=\linewidth]{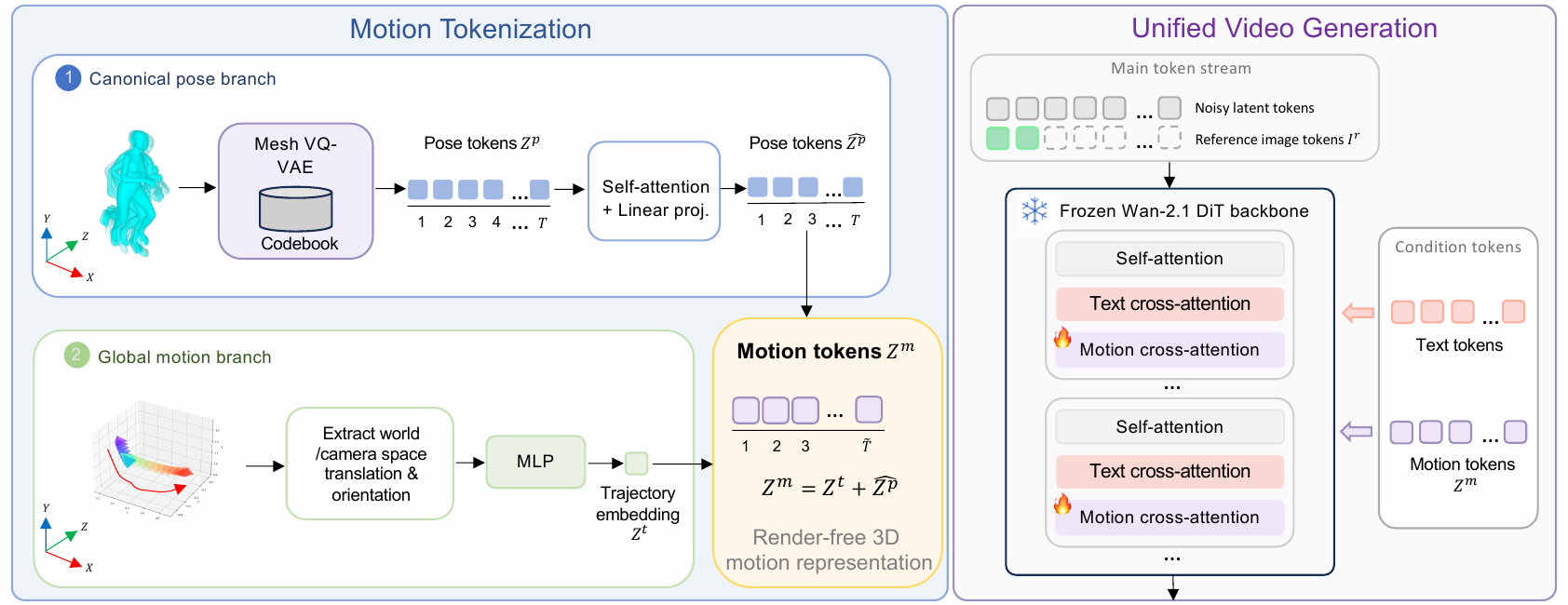}
    \caption{We explore the 3D awareness of video diffusion models through the task of human motion control. The figure illustrates the architecture of our proposed render-free human motion control framework with 3D human mesh tokenization. Given a reference image $I^{r}$, a motion sequence represented by 3D human meshes $\mathcal{M}_{1,\dots,T}$ (where $T$ is the number of frames), and the corresponding text prompt, we tokenize each modality into latent tokens using different tokenizers. Built on a Diffusion Transformer architecture, the reference image tokens (green) are concatenated with latent noise tokens, while the motion tokens (purple) and text tokens (pink) are injected into the video generation process through cross-attention. The framework operates without any intermediate 2D rendering.
    }
    \label{fig:pipeline}
\end{figure*}

\section{Related Work}
\subsection{Human Motion Control}
\paragraph{Motion Representation}
Representing human motion involves describing how a person moves over time. Early methods leverage optical flow~\cite{liang2024movideo, cong2023flatten, yan2023motion} to encode temporal dynamics between key frames and subsequent frames. However, optical flow suffers from ambiguities due to occlusion, mismatches, and newly exposed regions. Alternative approaches use depth sequences~\cite{esser2023structure, liang2024movideo} as motion representations, providing coarse 3D cues but lacking semantic structure. To exploit anatomical priors, many methods represent motion as 2D skeleton sequences~\cite{hu2024animate, wang2023disco, tan2024animate, zhou2024realisdance, peng2024controlnext, wang2024vividpose, zhang2024mimicmotion, men2024mimo, hu2025animate, wang2025unianimate, wang2025humandreamer, gan2025humandit}. While efficient and robust to appearance variation, skeletons discard surface geometry and cannot capture fine-grained deformations.

With advances in monocular human motion recovery~\cite{goel2023humans, shin2024wham, shen2024world}, recent works adopt parametric 3D human models, SMPL~\cite{loper2023smpl} and SMPL-X~\cite{pavlakos2019expressive}, as richer representations. These models describe the human body as a deformable triangular mesh with pose and shape parameters, offering vertex-level detail and geometric plausibility. To guide diffusion models, most existing methods render the human meshes into 2D guidance signals such as depth maps~\cite{zhu2024champ, hu2025animate, liang2025realismotion}, normal maps~\cite{zhu2024champ, liang2025realismotion}, semantic segmentation maps~\cite{zhu2024champ}, or color maps~\cite{zhou2024realisdance, zhou2025realisdance, liang2025realismotion}. In contrast, our method conditions video generation directly on 3D human mesh tokens: we compress the 3D human mesh into compact tokens without rendering, preserving full geometric fidelity and enabling more faithful 3D-aware motion control.

\vspace{-0.4cm}
\paragraph{Motion-Guided Video Generation}
To incorporate motion information into video generation models, existing methods either employ learnable encoders~\cite{zhu2024champ, hu2024animate, wang2023disco, jiang2025vace} or reuse frozen encoders from the base model (e.g., VAE)~\cite{men2024mimo, zhou2025realisdance, liang2025realismotion}. For instance, Champ~\cite{zhu2024champ} designs a dedicated encoder with convolutional and self-attention layers, while RealisMotion~\cite{liang2025realismotion} directly leverages the pre-trained Wan2.1 VAE~\cite{wang2025wan}. The extracted motion features are typically concatenated or added to the video features, as they are spatially aligned~\cite{wang2023disco, lin2025omnihuman}.

In addition to motion injection, how to transfer appearance from a reference image to the video is another important design choice. Since the reference may not align spatially with the target pose, attention-based mechanisms are widely adopted~\cite{vaswani2017transformer}. Animate Anyone~\cite{hu2024animate} proposes a ReferenceNet to extract multi-level features and inject them via cross-attention; this design has been extended in multiple subsequent works~\cite{hu2025animate, zhou2024realisdance, tong2024musepose, zhu2024champ, pang2025dreamdance, xuan2025rethink}. More recently, as video diffusion models shift toward Transformer architectures~\cite{rombach2022high, blattmann2023stable, wang2025wan}, more recent works compress the reference image into visual tokens and concatenate them with video tokens, enabling joint self-attention over reference image and target video~\cite{wang2025unianimate, gan2025humandit, zhou2025realisdance, liang2025realismotion, tan2025animate}.

\subsection{3D-Informed Video Generation}
Given that videos are 2D projections of underlying 3D scenes, recent works have attempted to integrate 3D priors into video diffusion models for better consistency and controllability~\cite{huang2025voyager, cao2025uni3c, fu20243dtrajmaster}. For example, some methods utilize the multi-view consistency of point clouds for camera control~\cite{cao2025uni3c, ren2025gen3c}, or 3D scene exploration~\cite{huang2025voyager, schneider2025worldexplorer, gu2025diffusion}. Another line of work represent foreground objects as simplified 3D primitives, such as 3D bounding boxes~\cite{wang2025cinemaster}, or spheres~\cite{chen2025perception}. Despite their diversity, nearly all existing methods rely on rendering 3D information into 2D feature maps before feeding them into the video diffusion backbone.

\section{Method}
\subsection{Overview}
In this paper, we study the 3D awareness of video diffusion models through the human motion control task, and propose a render-free framework that generates human motion videos conditioned on tokenized 3D human meshes. More specifically, as illustrated in Fig.~\ref{fig:pipeline}, our framework takes a reference image, a motion sequence, and the corresponding text prompt as inputs, and outputs a video in which the reference human moves according to the input motion. In the following, we first describe how to represent human motion and compress it into motion tokens. We then describe how to generate videos conditioned on the compressed motion tokens.

\subsection{Motion Representation and Tokenization}
\vspace{-0.1cm}
\paragraph{Motion Representation}
To represent human motion for video diffusion models, most existing methods adopt 2D projected motion representations, such as 2D skeletons. Although some methods use human depth maps or normal maps, these are still essentially 2.5D representations with projected and reduced information, and thus cannot fully capture 3D motion in the physical world. To model human motion more faithfully, we represent it as a sequence of 3D human meshes using the parametric human model SMPL~\cite{loper2023smpl}. Specifically, SMPL represents the human body as a function $\mathcal{F}(\gamma, \phi, \theta, \beta)$, parameterized by the global translation $\gamma\in\mathbb{R}^3$, global orientation $\phi\in\mathbb{R}^3$, body pose $\theta\in\mathbb{R}^{21\times3}$, and body shape $\beta\in\mathbb{R}^{10}$. With linear blend skinning and blend shape correction, SMPL outputs a 3D human mesh with 6,890 vertices and 13,776 faces. Human motion can therefore be naturally represented as a sequence of 3D human meshes. Compared with 2D or 2.5D motion representations, 3D human meshes provide richer priors on human shape, pose, and motion, capturing detailed and structured information about body geometry and spatial dynamics.

\vspace{-0.4cm}
\paragraph{Motion Tokenization}
For a video diffusion model, directly processing a human mesh $\mathcal{M}$ with vertex coordinates $\mathcal{V}\in\mathbb{R}^{6890\times 3}$ is challenging. Therefore, inspired by~\cite{fiche2024vq}, we compress the human mesh into discrete latent tokens using a Vector Quantized Variational Auto-Encoder (VQ-VAE)~\cite{van2017neural}. Specifically, to focus on modeling pose and avoid ambiguity during mesh encoding, we adopt a canonical mesh representation by translating the mesh to the origin and removing its global orientation. We also discard the body shape parameters during SMPL instantiation, as they are largely irrelevant to human motion. We then feed the canonical mesh into a fully convolutional mesh autoencoder~\cite{zhou2020fully} to map the input mesh vertices $\mathcal{V}$ to $N$ low-dimensional latent vectors that preserve the spatial structure of the human body. Each latent vector $z_n\in\mathbb{R}^L$ ($n=1,\dots,N$) corresponds to a localized body region, and is further quantized by assigning it to its nearest entry in a learned codebook of shape $S\times L$, where $S$ is the codebook size and $L$ is the latent dimension. This discrete representation is significantly more compact than the original vertex coordinates, while still capturing realistic human pose. In this way, a canonical 3D human mesh is represented as a group of $N$ discrete latent vectors $\{z_1,\dots,z_N\}$, abbreviated as $Z^{p}\in\mathbb{R}^{N\times L}$:
\begin{align}
    Z^{p}=\{z_1,\dots,z_N\}=Q(\phi(\mathcal{V})),
\end{align}
where $Q$ denotes the vector quantization process and $\phi$ is the encoder of the mesh autoencoder. Note that $z_n$ is a quantized latent embedding looked up from the codebook.

For a video sequence, we first compress each mesh $\mathcal{M}_t$ ($t=1,\dots,T$) individually into pose latents, yielding a sequence $\{Z^p_1,\dots,Z^p_T\}$. Since the latent video representation in video diffusion models often has a lower temporal resolution than the input video, we temporally reshape the pose sequence before feature projection to align it with the latent video frames used in the next subsection. We then apply several one-dimensional self-attention ($\text{SA}$) blocks along the sequence dimension with a residual connection to capture dependencies among latent vectors and improve the coherence of motion across body parts. The resulting features are projected to the transformer channel dimension $D$ used by the video diffusion model, yielding the pose token group $\hat{Z}^{p}\in\mathbb{R}^{N\times D}$:
\begin{align}
    \hat{Z}^{p} &= \text{Proj}(Z^{p} + \text{SA}(Z^{p})),
\end{align}
where $\text{Proj}$ is a linear projection layer. For simplicity, the temporal reshaping operation is omitted.

Since the canonical mesh removes global translation and orientation, we reintroduce them as separate global motion cues by injecting them into the pose token group. We represent the global translation as a vector $\gamma\in\mathbb{R}^3$ and the global orientation as a rotation matrix $R\in\mathbb{R}^{3\times 3}$. To additionally encode camera-related information, we use translation and orientation cues from both world space and camera space. The world space cues capture view-independent global motion, while camera space cues provide view-dependent motion information aligned with image observations. For a motion sequence, we reshape the temporal dimension of these global motion cues into the channel dimension, concatenate them, and then project them with an MLP into a trajectory embedding $Z^{t}\in\mathbb{R}^{D}$:
\begin{align}
    Z^{t}=\text{MLP}([\gamma_{w},\text{vec}(R_{w}),\gamma_{c},\text{vec}(R_{c})]),
\end{align}
where $\gamma_{w}$ and $R_{w}$ denote the global translation and orientation in world space, while $\gamma_{c}$ and $R_{c}$ denote those in camera space. The function $\text{MLP}(\cdot)$ denotes a multi-layer perceptron for feature projection. For simplicity, we omit the temporal reshaping operator in the equation.

We then broadcast the trajectory embedding $Z^{t}$ to all pose tokens and add it to the pose token group $\hat{Z}^{p}$ to obtain the final motion token group $Z^{m}\in\mathbb{R}^{N\times D}$:
\begin{align}
    Z^{m}=Z^{t} + \hat{Z}^{p}. 
\end{align}

In practice, for a sequence of human meshes $\mathcal{M}_{1,\dots,T}$ in a $T$-frame video, we obtain a sequence of motion tokens $\{Z^{m}_1,\dots,Z^{m}_\frac{T}{s}\}$ (abbreviated as $Z^{m}_{1,\dots,\frac{T}{s}}\in\mathbb{R}^{\frac{T}{s}\times N\times D}$; $s$ is the temporal downsampling factor) after temporal alignment with the latent video representation. Each motion token group has $N$ tokens and corresponds to one latent video frame in the video diffusion model.

By representing human motion as tokens instead of rendered guidance, we preserve 3D geometric information and fine-grained motion details while minimizing information loss. Compared with directly using mesh vertices or SMPL parameters, this representation is much more compact and easier for video diffusion models to process, while still retaining global translation, global orientation, and detailed 3D structural information of the human body.

\subsection{Unified Human Motion Control}
After compressing 3D human motion into a sequence of motion tokens, we then describe how to inject these structured 3D motion tokens into a pretrained video diffusion model for controllable human video generation. Our goal is to enable the model to better capture 3D human motion and achieve precise control over human dynamics, while fully leveraging the strong appearance and motion priors of the pretrained backbone. To this end, we build our framework upon the Wan-2.1 image-to-video (I2V) model~\cite{wang2025wan} and augment it with motion-conditioning modules.

As illustrated in Fig.~\ref{fig:pipeline}, we freeze the parameters of the Wan-2.1 I2V model and inject the motion tokens into the DiT backbone through per-latent-frame cross-attention modules, as each motion token group corresponds to one latent video frame. The motion cross-attention block is inserted after every text cross-attention block. More formally, given the visual token feature $Z_{t,l}^v\in\mathbb{R}^{M\times D}$ at the $t$-th latent video frame after text cross-attention in the $l$-th DiT block, we further condition it on the corresponding motion tokens $Z_t^m\in\mathbb{R}^{N\times D}$ via cross-attention:
\begin{align}
\bar{Z}_{t,l}^v=Z_{t,l}^v+\text{CA}(Z_{t,l}^v, Z_t^m),
\end{align}
where $\bar{Z}_{t,l}$ is the output and is then fed into the $(l+1)$-th DiT block. This design explicitly enforces temporal alignment between the latent video frames and motion tokens, allowing each latent frame to attend only to the motion tokens at the corresponding time step. It also reduces the computational cost compared with global cross-attention over all motion tokens.

During training, we initialize the last projection layer of each per-latent-frame cross-attention module to zero to avoid disrupting the pretrained backbone at the beginning of finetuning. We then finetune only the newly added modules under the rectified flow matching objective~\cite{lipman2022flow, esser2024scaling}, while freezing the original model parameters as well as the motion tokenizer. During inference, we use DDIM~\cite{song2020denoising} for accelerated sampling.

By tokenizing human meshes into motion tokens, our framework provides a simple and unified interface for motion representation and control. All modalities, including image, video, text, and motion, are represented as tokens and integrated within a unified token-based framework. In addition, our framework naturally supports disentangled control over trajectory, orientation, and pose~\cite{liang2025realismotion} thanks to our decomposed human motion representation.

\section{Experiments}
\subsection{Experimental Setup}
\vspace{-0.1cm}
\paragraph{Architecture}
For motion tokenization, we use an MLP block with two linear layers and a GELU activation in between to obtain the trajectory embedding $Z^t\in\mathbb{R}^{5120}$. Following Wan-2.1 I2V, the latent video representation has a temporal downsampling factor of $s=4$, so a 97-frame input sequence is aligned to $25$ latent video frames. For pose representation, we follow the architecture of the fully convolutional mesh autoencoder~\cite{zhou2020fully} and use the pretrained weights in~\cite{fiche2024vq} to compress the canonical human mesh vertices $\mathcal{V}\in\mathbb{R}^{6890\times 3}$ into $Z^p\in\mathbb{R}^{54\times 9}$. In other words, the human pose is represented as a group of 54 tokens, each with 9 dimensions. We first project the pose tokens from 9 to 512 dimensions, process them with 5 self-attention blocks with 16 heads, and then project the resulting features to the hidden dimension of the DiT model (5120). For unified human motion control, we build our model upon the open-source video diffusion model Wan-2.1 I2V~\cite{wang2025wan}. For the inserted motion cross-attention blocks, we use the same configuration as the text cross-attention blocks. The final model contains 20.33B parameters.

\vspace{-0.4cm}
\paragraph{Training}
We train our model on an internal human motion dataset containing about 300K video clips with multiple resolutions and frame rates. For each video, we estimate human meshes using GVHMR~\cite{shen2024world} and obtain the corresponding text captions using LLaVA~\cite{liu2024llavanext}. Since we only train the newly added motion modules that are independent of spatial resolution, we uniformly sample 97 frames at 16 FPS and choose the first frame as the reference image. Moreover, we adopt a cascaded training strategy to gradually increase the spatial resolution: we first train on 240p for 160,000 iterations with a batch size of 64, then on 480p for 80,000 iterations with a batch size of 32, and finally on 720p for 40,000 iterations with a batch size of 8. We use the AdamW optimizer~\cite{loshchilov2017adamw} for optimization, and the full training process takes about 20 days. The learning rate and weight decay are set as 1e-5 and 1e-4, respectively. During training, to improve the model robustness during motion editing as well as prepare for classifier-free guidance in inference, we randomly drop conditioning inputs with a probability of 0.1 by setting the text prompt to a fixed negative prompt and setting the trajectory representation and pose tokens to zero. Other settings, such as the noise scheduler, are kept the same as in Wan-2.1 I2V.

\vspace{-0.4cm}
\paragraph{Inference and Evaluation}
During inference, we use classifier-free guidance with a scale of 5.0 and perform DDIM sampling with 40 steps. It takes about 4 minutes to generate a 97-frame 720p video on 8 H100 GPUs. For evaluation, we use VBench-I2V metrics~\cite{huang2023vbench}, PSNR, SSIM, LPIPS, FID, FVD, Translation Error, and Rotation Error for quantitative comparison on benchmark datasets.

\begin{table*}[t]
\captionsetup{font=small}%
\small
\caption{Comparison with existing methods on the Trajectory100 dataset~\cite{liang2025realismotion} with global translation and rotation evaluation metrics.}
\vspace{-0.5cm}
\label{tab:trajectory100}
\begin{center}
\begin{tabular}{lccccccc}
  \toprule
  \scalebox{1}{Method}   &  \scalebox{1}{\makecell{Translation Error (m)$\downarrow$}} & \scalebox{1}{\makecell{Rotation Error (deg) $\downarrow$}} & \scalebox{1}{PSNR$\uparrow$} & \scalebox{1}{SSIM$\uparrow$} & \scalebox{1}{LPIPS$\downarrow$}& \scalebox{1}{FID$\downarrow$}  & \scalebox{1}{FVD$\downarrow$} \\ \midrule
  Wan-2.1-I2V~\cite{wang2025wan} & 10.349 & 0.418 & 14.96 & 0.4763 & 0.3260 & 33.06 & 1421.87  \\
  Tora~\cite{zhang2024tora} & 5.667 & 0.355 & \underline{16.56} & \underline{0.5195} & 0.2501 & \textbf{21.51} & 957.81\\
  RealisDance-DiT~\cite{zhou2025realisdance} & \underline{1.706} & \textbf{0.167} & 16.17 & 0.4892 & \underline{0.2481} & {23.02}  & \underline{758.08}\\
  \color{gray}{RealisMotion$^\star$} & \color{gray}{1.198} & \color{gray}{0.101} & \color{gray}{22.57} & \color{gray}\color{gray}{0.7664} & \color{gray}{0.0686} & \color{gray}{12.00} & \color{gray}{314.59}\\
  \midrule
  \textbf{MeshToken} (ours) & \textbf{1.697} & \underline{0.173} & \textbf{16.78} & \textbf{0.5245} & \textbf{0.2451} & \underline{22.13} & \textbf{695.62}\\
  
  \bottomrule
\end{tabular}
\end{center}
\centering
\end{table*}

\begin{table*}[t!]
\centering
\captionsetup{font=small}
\small
\setlength{\tabcolsep}{5.5pt}
\caption{Comparison with existing methods on the RealisDance-Val~\cite{zhou2025realisdance} dataset.}
\vspace{-0.2cm}
\label{tab:realisdancedit}
\begin{tabular}{lcccccccc}
\hline
Method 
& \makecell{I2V\\Subject $\uparrow$} 
& \makecell{I2V\\BG $\uparrow$} 
& \makecell{Subject\\Consist. $\uparrow$} 
& \makecell{Background\\Consist. $\uparrow$} 
& \makecell{Temporal\\Flicker $\uparrow$} 
& \makecell{Motion\\Smooth $\uparrow$} 
& \makecell{Dynamic\\Degree $\uparrow$} 
& \makecell{Aesthetic\\Quality $\uparrow$} \\ %
\hline
 Animate-X~\cite{tan2024animate}   & \textbf{96.06} & \underline{96.59} & 93.83 & 94.83 & 97.40 & 98.52 & 53 & 55.22 \\ %
    ControlNeXt~\cite{peng2024controlnext} & 92.91 & 93.92 & 91.41 & 93.57 & 96.91 & 98.05 & 63 & 55.57 \\ %
    MimicMotion~\cite{zhang2024mimicmotion} & 92.79 & 93.80 & 91.10 & 93.20 & 96.78 & 98.20 & 59 & 53.31 \\ %
    MooreAA~\cite{hu2024animate}     & 92.33 & 93.35 & 93.12 & 93.77 & 95.20 & 96.74 & \underline{68} & 56.08 \\ %
    MusePose~\cite{tong2024musepose}    & 92.24 & 93.01 & {93.88} & {94.88} & \underline{97.88} & {98.57} & 57 & {56.28} \\ %
    RealisDance-DiT~\cite{zhou2025realisdance} & {95.97} & {96.57} & \underline{93.91} & \underline{95.83} & {97.76} & \underline{98.71} & {66} & \underline{57.93} \\ %
    \midrule
    \textbf{MeshToken} (Ours)                  & \underline{96.03} & \textbf{97.18} & \textbf{94.04} & \textbf{96.23} & \textbf{98.32} & \textbf{98.79} & \textbf{68} & \textbf{58.14} \\ %
\hline
\end{tabular}
\end{table*}

\begin{figure}[t]
\captionsetup{font=small}%
\small
\begin{center}
\includegraphics[width=0.48\textwidth, page=2]{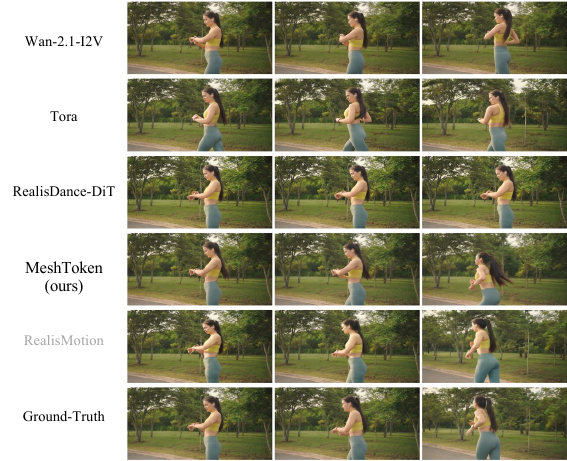}
\caption{Visual comparison with existing methods on the Trajectory100 dataset. Our method generates videos with better trajectories and body poses. Note that RealisMotion takes the video background as input. Please zoom in for a clearer view of the differences.}
\label{fig:action1}
\end{center}
\vspace{-0.7cm}
\end{figure}

\begin{figure}[t]
\captionsetup{font=small}%
\small
\begin{center}
\includegraphics[width=0.48\textwidth, page=1]{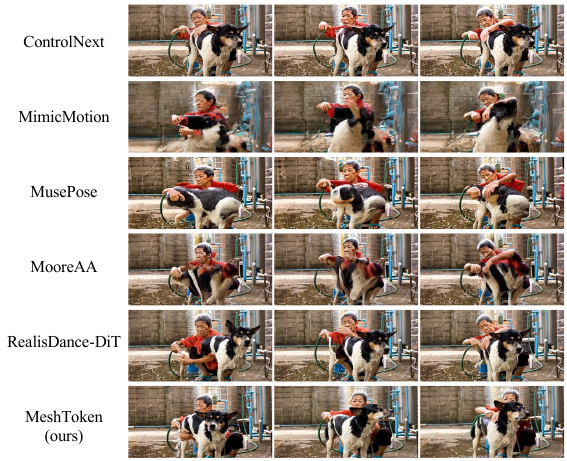}
\caption{Visual comparison with existing methods on the RealisDance-Val dataset. Our method can better deal with human-object interaction and occlusion. Please zoom in for a clearer view of the differences.}
\label{fig:action2}
\end{center}
\vspace{-0.7cm}
\end{figure}

\begin{table*}[t]
\centering
\captionsetup{font=small}
\small
\setlength{\tabcolsep}{4.6pt}
\caption{Comparison on videos with edited motions. We edit the original actions as ``Walking'' and ``Jogging'', respectively.}
\vspace{-0.2cm}
\label{tab:vbench_i2v}
\begin{tabular}{llcccccccc}
\hline
Category & Method 
& \makecell{I2V\\Subject $\uparrow$} 
& \makecell{I2V\\BG $\uparrow$} 
& \makecell{Subject\\Consist. $\uparrow$} 
& \makecell{Background\\Consist. $\uparrow$} 
& \makecell{Temporal\\Flicker $\uparrow$} 
& \makecell{Motion\\Smooth $\uparrow$} 
& \makecell{Dynamic\\Degree $\uparrow$} 
& \makecell{Aesthetic\\Quality $\uparrow$} \\
\hline
\multirow{3}{*}{Walking}
& Wan2.1-I2V~\cite{wang2025wan}              & 96.02 & 96.21 & 93.52 & 93.74 & 96.94 & 98.65 & 61 & \textbf{56.33} \\
& RealisDance-DiT~\cite{zhou2025realisdance} & \textbf{96.09} & 96.37 & 93.69 & 94.42 & 96.31 & 97.88 & \textbf{63} & 55.77 \\
& \textbf{MeshToken} (Ours)                  & {96.03} & \textbf{96.86} & \textbf{93.71} & \textbf{95.79} & \textbf{97.32} & \textbf{98.79} & 60 & {56.12} \\
\hline
\multirow{3}{*}{Jogging}
& Wan2.1-I2V~\cite{wang2025wan}              & 95.94 & 96.02 & \textbf{93.66} & 93.19 & 96.46 & 97.31 & 65 & 56.32 \\
& RealisDance-DiT~\cite{zhou2025realisdance} & 96.01 & 96.18 & 93.46 & 93.87 & 95.85 & 96.67 & \textbf{69} & 54.01 \\
& \textbf{MeshToken} (Ours)                  & \textbf{96.11} & \textbf{96.63} & {93.41} & \textbf{95.16} & \textbf{97.58} & \textbf{98.34} & {68} & \textbf{56.48} \\
\hline
\end{tabular}
\end{table*}

\subsection{Results on Benchmark Datasets}
We evaluate human motion control on two representative benchmarks, Trajectory100~\cite{liang2025realismotion} and RealisDance-Val~\cite{zhou2025realisdance}. Before assessing video generation quality, we first measure the average per-vertex reconstruction error of the recovered human meshes on these datasets. The error is 8.0 mm on Trajectory100 and 7.1 mm on RealisDance-Val. These low reconstruction errors indicate that the human meshes are faithfully compressed and that most of the original motion information is preserved during motion tokenization.

We then compare the video generation results in Table~\ref{tab:trajectory100} and Table~\ref{tab:realisdancedit}. As shown in Table~\ref{tab:trajectory100}, our method achieves strong overall performance on most fidelity and perceptual quality metrics. We note that RealisMotion~\cite{liang2025realismotion} leverages the video background as an additional input signal; therefore, we do not include it in the direct comparison and mark it with a $^\star$. In addition, this table also reports trajectory controllability evaluated by estimating human motion with GVHMR and comparing it with the target motion. Our method obtains Translation Error and Rotation Error comparable to those of the most closely related baseline, RealisDance-DiT, which adopts the same video diffusion backbone but relies on rendered 2D motion images for control. This suggests that our 3D motion representation can achieve competitive controllability without requiring explicit 2D rendering-based guidance. As shown in Table~\ref{tab:realisdancedit}, our method achieves strong overall performance on the RealisDance-Val benchmark, attaining the best results on most metrics. This further confirms the effectiveness of our method.

We provide visual comparisons in Fig.~\ref{fig:action1} and Fig.~\ref{fig:action2}. In Fig.~\ref{fig:action1}, even under substantial camera viewpoint changes, our model still generates the correct action, whereas competing methods may produce incorrect motions. We attribute this robustness to our 3D motion representation, which naturally generalizes across different viewpoints. In Fig.~\ref{fig:action2}, our model better handles object occlusion and captures complex human-object interactions. In contrast, competing methods often generate incorrect actions, exhibit severe artifacts, or produce unnatural videos in these challenging cases. For example, a human hand may incorrectly pass through the dog's body. We attribute the advantage of our method to its ability to more fully exploit the 3D reasoning capability of the video diffusion model, rather than restricting motion control to rendered 2D motion signals.

\subsection{Results on Edited Motions}
Unlike motions extracted from natural videos, edited motions often involve imperfect combinations of trajectories and body poses, and may not align well with the background. Previous methods are typically guided by rendered 2D condition maps, which might generate unrealistic videos under unrealistic motion guidance. In contrast, our method decomposes motion into trajectory and body pose and feeds them into the model separately, without rendering. We argue that this design can help improve the model's robustness to imperfect motion editing and bridge the gap between natural training data and the imperfect combinations of trajectory and body pose encountered during editing.

\begin{figure}[t]
\captionsetup{font=small}%
\small
\begin{center}
\includegraphics[width=0.48\textwidth, page=1]{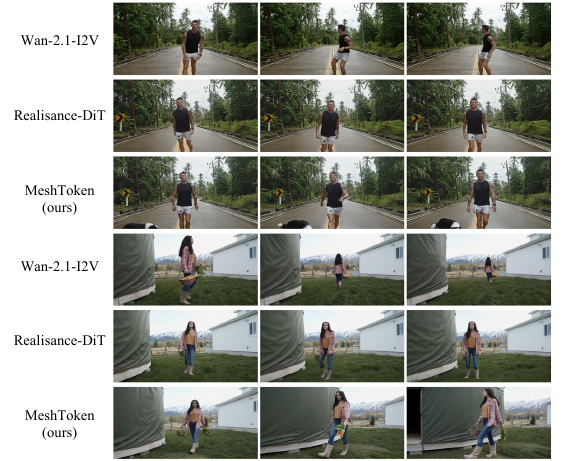}
\caption{Visual comparison with existing methods on motion editing. Our method can generate natural videos under imperfect combination of trajectories and body poses. Please zoom in for a clearer view of the differences.}
\label{fig:action_edit}
\end{center}
\end{figure}

To validate this claim, we select 50 videos from the Trajectory100 dataset~\cite{liang2025realismotion} and edit their actions as two representative motion patterns: ``Walking'' and ``Jogging'', from the action bank introduced in~\cite{liang2025realismotion}. As shown in Table~\ref{tab:vbench_i2v}, compared with the baseline model RealisDance-DiT~\cite{zhou2025realisdance} built on the same base model, our method achieves better performance in temporal flickering and motion smoothness. This suggests that our approach generates smoother motions and reduces temporal instability from edited motions. In addition, it also improves background consistency, possibly because the model learns to infer camera-view changes implicitly during training. These improvements are further supported by Fig.~\ref{fig:action_edit}, where our method produces more natural movements with fewer artifacts. In contrast, RealisDance-DiT tends to follow the imperfect edited motion more rigidly, resulting in unnatural outputs, while Wan-2.1-I2V cannot control the trajectory and body pose accurately.

\subsection{Ablation Study}
\begin{figure}[t]
\captionsetup{font=small}%
\small
\begin{center}
\includegraphics[width=0.48\textwidth, page=1]{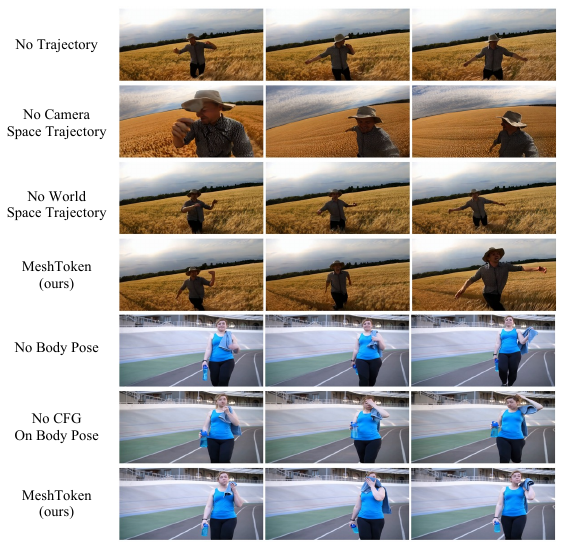}
\caption{Visual comparison for the ablation study. Removing trajectory inputs leads to severe camera shake or overly stationary motion, while removing body pose guidance degrades fine-grained motion quality and action accuracy. Please zoom in for a clearer view of the differences.}
\label{fig:ablation_study_visual}
\end{center}
\vspace{-0.3cm}
\end{figure}

\vspace{-0.1cm}
\paragraph{Impact of Trajectory}
To study the role of trajectory, we ablate the trajectory condition by fully or partially zeroing it out. As shown in Table~\ref{tab:ablation_study}, removing the trajectory entirely leads to a significant performance drop, as the model can only rely on the text prompt, which typically lacks precise spatial information. Removing either the camera-space or world-space trajectory also degrades performance, indicating that both are important for modeling global motion and viewpoint changes. For example, as shown in Fig.~\ref{fig:ablation_study_visual}, removing the full trajectory leads to incorrect motions, while removing either component causes severe camera shake or overly stationary motion when the world-space or camera-space trajectory is omitted. In addition, we find that classifier-free guidance on the trajectory condition improves trajectory controllability during inference.

\vspace{-0.4cm}
\paragraph{Impact of Body Pose}
We further investigate the role of body pose by removing the pose input from the model. As shown in Table~\ref{tab:ablation_study}, this results in clear performance degradation, as the model no longer receives detailed action guidance. We also ablate the pose refinement blocks and observe an additional drop in performance. We attribute this to the fact that different motion tokens correspond to different body parts~\cite{fiche2024vq}; the attention layers in the refinement blocks help improve motion coherence across body parts and over time. Similar to the trajectory condition, we find that applying classifier-free guidance to the pose tokens during inference further improves performance. We provide visual examples in Fig.~\ref{fig:ablation_study_visual}. As shown in the figure, removing the body pose input causes the model to fail to generate the correct action of wiping the face with a towel, while removing classifier-free guidance leads to errors in modeling the interaction between the hand and the towel, causing the model to generate a hand-wiping motion instead.

\begin{table}[t]
\centering
\captionsetup{font=small}%
\small
\caption{Ablation study on different design choices.} %
\begin{tabular}{lcc}
    \toprule
    Ablation Study & PSNR$\uparrow$ & LPIPS$\downarrow$ \\
    \midrule
    No Trajectory Input & 15.32 & 0.2941 \\
    Camera-Space Trajectory Only & 16.17 & 0.2781 \\
    World-Space Trajectory Only & 16.32 & 0.2688 \\
    w/o CFG on Trajectory & 16.31 & 0.2521 \\
    \midrule
    No Pose Input & 15.61 & 0.3131 \\
    No Pose Refinement & 16.38 & 0.2520 \\
    w/o CFG on Pose & 16.19 & 0.2548 \\
    \midrule
    MLP for Mesh Tokenization  & 14.79 & 0.3271 \\
    SMPL Parameters as Motion  & 15.91 & 0.2899 \\
    \midrule
    No Per-Frame Cross Attention  & 16.49 & 0.2591 \\
    Unfreeze the Video Backbone  & 16.81 & 0.2452 \\
    \midrule
    \textbf{MeshToken} (ours) & \textbf{16.78} & \textbf{0.2451}\\
    \bottomrule
\end{tabular}
\label{tab:ablation_study}
\end{table}

\vspace{-0.4cm}
\paragraph{Impact of Motion Representation}
In addition to rendered 2D motion guidance and our compressed motion tokens, other motion representations are also possible. We test two alternatives: using an MLP to directly encode 3D meshes, or directly using SMPL parameters as motion input. As shown in Table~\ref{tab:ablation_study}, both perform worse than our method, suggesting that they are less effective for motion control. We conjecture that these representations are either too unstructured or too difficult for the video diffusion model to exploit effectively.

\vspace{-0.4cm}
\paragraph{Impact of Per-Frame Cross-Attention}
We inject motion guidance into the video diffusion model through per-frame cross-attention, since motion tokens are temporally aligned with video frames and differ from text tokens. To validate this design, we replace the per-frame cross-attention with standard cross-attention, which allows each video frame to attend to motion tokens from all time steps. As shown in Table~\ref{tab:ablation_study}, this change leads to a performance drop. One possible reason is that, without explicit temporal alignment, the model may confuse motion signals from different frames.

\vspace{-0.4cm}
\paragraph{Impact of Freezing the Video Backbone}
We compare two training strategies: freezing the video diffusion backbone and updating only the newly added motion cross-attention blocks, or finetuning the entire model jointly. As shown in Table~\ref{tab:ablation_study}, the two strategies yield similar results. Therefore, to reduce training cost and better isolate the 3D understanding capability of the pretrained video diffusion model, we freeze the video backbone in our final design.

\section{Conclusion}
In this paper, we study the 3D awareness of video diffusion models through human motion control. We propose a render-free framework that directly conditions video generation on tokenized 3D human meshes, avoiding the information loss and view-dependent bias of rendering-based motion guidance. By decomposing motion into trajectory and body pose and representing them as compact tokens, our method provides a faithful and flexible interface for controllable video generation. Experiments on benchmark datasets and edited-motion settings show that our method achieves strong performance and better robustness to viewpoint changes and imperfect motion editing. These results suggest that video diffusion models can capture richer 3D structure when equipped with appropriate 3D motion representations. 

In future work, we plan to extend this render-free tokenization paradigm to more general 3D objects and dynamic scenes. We believe this direction could enable more direct 3D-aware video generation without intermediate rendering.

{\small
\bibliographystyle{ieee_fullname}
\bibliography{superresolution.bib}
}

\end{document}